\documentclass[final]{cvpr}

\usepackage[dvipsnames]{xcolor}
\usepackage{times}
\usepackage{epsfig}
\usepackage{graphicx}
\usepackage{amsmath}
\usepackage{amssymb}
\usepackage{subcaption}
\newcommand{\myparagraph}[1]{\vspace{2pt}\noindent\textbf{#1}}

\newcommand{\real}{{\rm I\!R}}
\newcommand{\set}{\mathcal}
\newcommand{\ours}{PAnS}
\newcommand{\expandednick}{Prototypical Anomaly Segmentation}

\usepackage[pagebackref=true,breaklinks=true,colorlinks,bookmarks=false]{hyperref}

\def\cvprPaperID{26} % *** Enter the CVPR Paper ID here
\def\confYear{SAIAD 2021}

\begin{document}

%%%%%%%%% TITLE
\title{Detecting Anomalies in Semantic Segmentation with Prototypes
\vspace{-14pt}}
\author{
Dario Fontanel$^{1}$, Fabio Cermelli$^{1,2}$, Massimiliano Mancini$^{3}$, Barbara Caputo$^{1,2}$\\
$^1$Politecnico di Torino, $^2$Italian Institute of Technology, $^3$University of Tübingen \\
{\tt\small \{dario.fontanel, fabio.cermelli, barbara.caputo\}@polito.it,} \\
{\tt\small massimiliano.mancini@uni-tuebingen.de}

\vspace{-14pt}
}

\maketitle

%%%%%%%%% ABSTRACT
\begin{abstract}
\vspace{-10pt}
Traditional semantic segmentation methods can recognize at test time only the classes that are present in the training set. This is a  significant limitation, especially for semantic segmentation algorithms mounted on intelligent autonomous systems, deployed in realistic settings. Regardless of how many classes the system has seen at training time, it is inevitable that unexpected, unknown objects will appear at test time. The failure in identifying such anomalies may lead to incorrect, even dangerous behaviors of the autonomous agent equipped with such segmentation model when deployed in the real world. Current state of the art of anomaly segmentation uses generative models, exploiting their incapability to reconstruct patterns unseen during training. However, training these models is expensive, and their generated artifacts may create false anomalies. 
In this paper we take a different route and we propose to address anomaly segmentation through prototype learning.
Our intuition is that anomalous pixels are those that are dissimilar to all class prototypes known by the model.
We extract class prototypes from the training data in a lightweight manner using a cosine similarity-based classifier. Experiments on StreetHazards show that our approach achieves the new state of the art, with a significant margin over previous works, despite the reduced computational overhead. Code is available at \url{https://github.com/DarioFontanel/PAnS}.

%assume that at test time only the classes that have been seen during training will be encountered. Clearly this is a significant limitation which in many situations can lead to an incorrect and dangerous behavior of the model when deployed in the real world. 
%In this paper, we address the more realistic scenario in which at test time it is possible to face unknown classes. Although state of the art methods focus only on recognizing these anomalies, in our work we emphasize how not confusing segmentation boundaries as anomalies is just as important. To prove our hypothesis, in our framework we propose a novel technique to increase the confidence of the model on segmentation boundaries and to increase the model entropy on anomaly pixels. We show that our approach is superior to the state of the art on all benchmarks currently used.
\end{abstract}

%%%%%%%%% BODY TEXT\section{Introduction}
\vspace{-15pt}
\section{Introduction}
\vspace{-5pt}
\label{sec:introduction}
\begin{figure}[tb]
  \centering
  \includegraphics[width=\columnwidth]{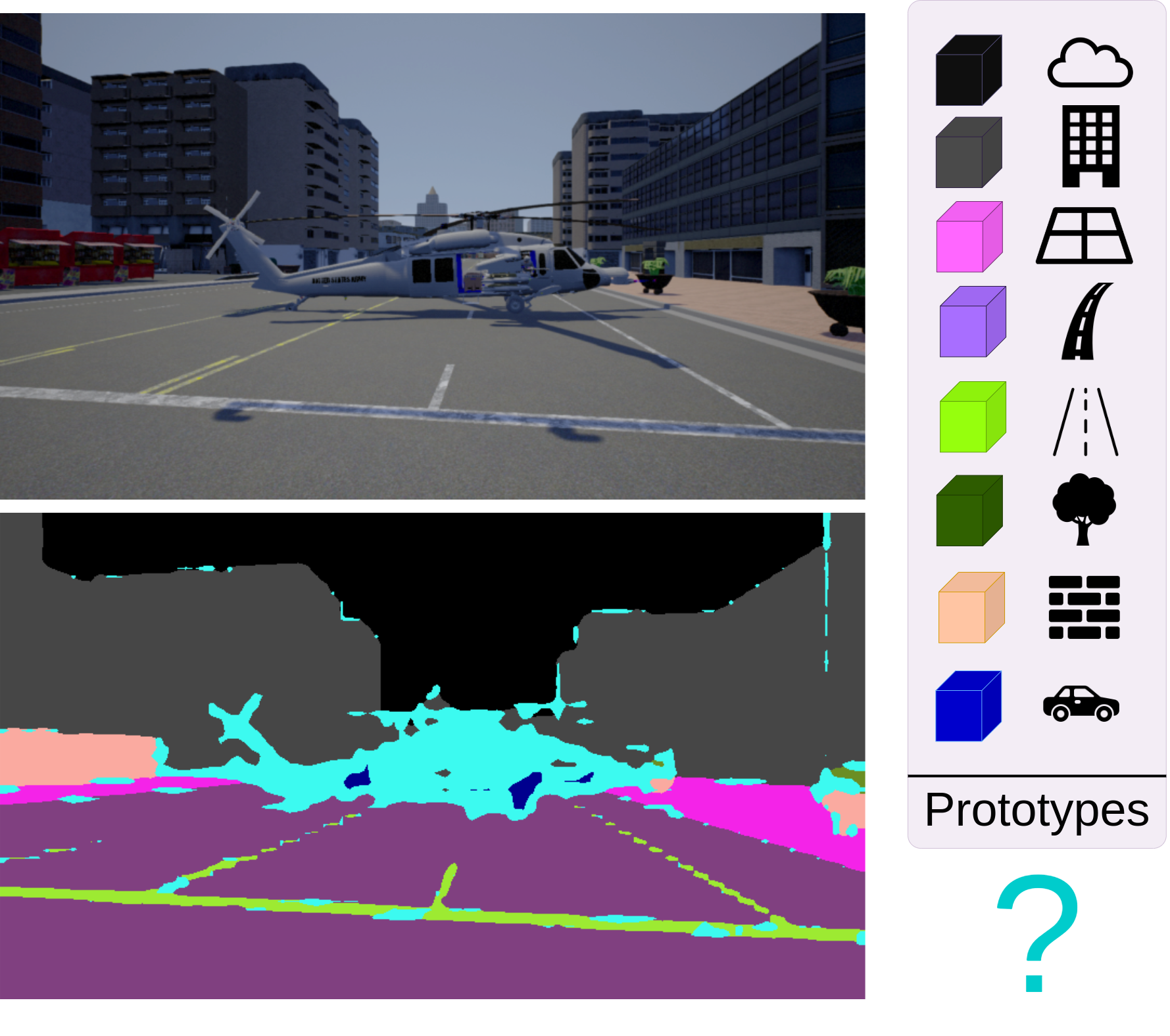} 
   \vspace{-20pt}
   \caption{Anomaly segmentation (AS) aims to segment objects unseen to the model. Addressing AS is crucial, especially in autonomous driving applications, where confusing an anomalous object with known ones can be extremely dangerous. In this work, we address AS through prototype learning, where the anomalies (light-blue) are all regions unmatched with any class prototype learned by the model.}
  \vspace{-20pt}
  \label{fig:teaser}
 \end{figure}

For machines acting in the real world, it is of the utmost importance recognizing what objects are present in their surroundings,  and where. To achieve this objective, multiple works have focused on the task of semantic segmentation \cite{long2015fully,chen2017deeplab}, where the goal is to assign to each pixel in the image its corresponding semantic label. However, semantic segmentation models are inherently limited by the classes they see annotated at training time. As vast as their training database might be, this limitation is a crucial weakness, % for any visual system applied in the real world, 
as clearly it is not possible to capture in a static collection all the possible semantic classes a system might ever encounter.  Ideally, we would like a segmentation model to recognize whether a pixel belongs to one of its known classes, or whether it belongs to an unseen category not included in the training set. Note that this capability is important whenever pixels of unknown categories might be a threat for the machine (or the human) making use of a semantic segmentation module. As an example, since no semantic segmentation dataset contains labeled pixels for the class \textit{helicopter}, if an autonomous driving system faces a scene as the one depicted in Figure~\ref{fig:teaser} it will have no chances to avoid a fatal collision, unless it recognizes that there is something unforeseen in the image, \ie an \textit{anomaly}.
% Variability of real world, necessity to recognize what is unknown, reference to figure. 

In this work, we focus on the problem depicted in Figure~\ref{fig:teaser}, namely anomaly segmentation (AS) \cite{haselmann2018anomaly,xia2020synthesize}. The goal of AS is to recognize whether a pixel in an image belongs to a category unseen during training, being unknown to the model. Previous works addressed this problem by either imposing a threshold on the predicted probabilities per pixel \cite{hendrycks2016baseline} or by means of generative approaches, 
%comparing input (or features) images with a reconstructed version of the same 
comparing input images (or features) with their reconstructed versions %of the respective ones}
\cite{lis2019detecting,xia2020synthesize}. However, both strategies present some drawbacks. {
%The first ignores the fact that the predicted probabilities \textit{after normalization} do not contain information about the original score given to each class (\ie two classes predicted with high confidence have equal and low probability after the softmax operation). 
The first ignores that the softmax function blurs %the information of 
the confidence of the model regarding the presence of a certain class, %. This information resides in the %difference between the 
%original scores  
\ie after the softmax, two classes predicted with high scores (logits) have nevertheless equal and low probability).
}

\begin{figure}[t]
    \centering
    \includegraphics[width=0.8\linewidth]{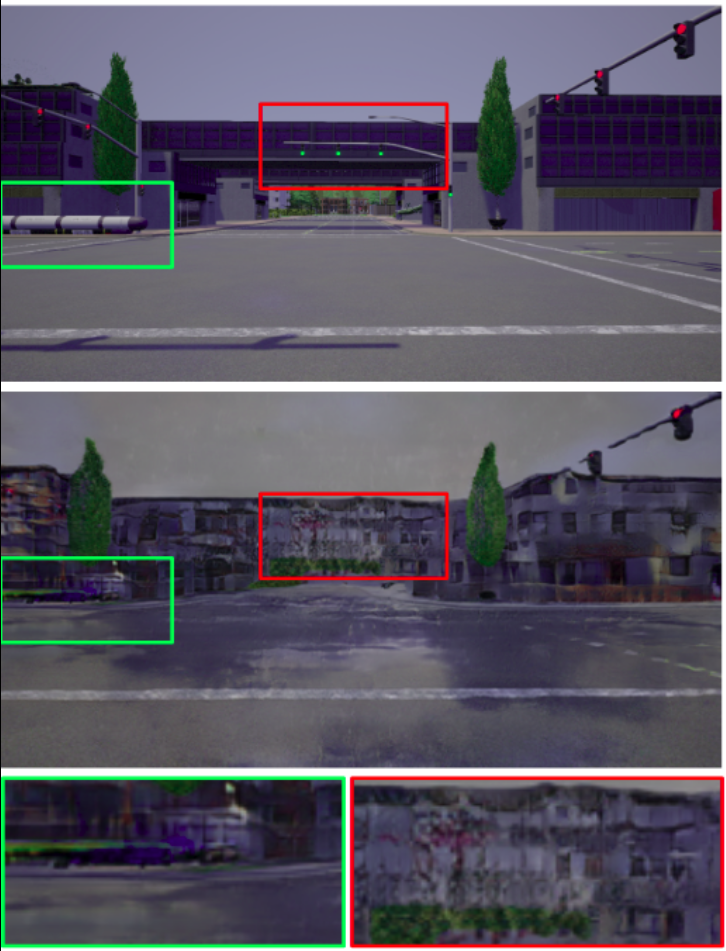}
     \caption{Qualitative results of SPADE \cite{park2019semantic} reconstructions on an image from StreetHazards dataset \cite{hendrycks2019benchmark} (top). The green box shows the anomaly, correctly not reconstructed by the model. The red box, instead, shows one of the artifacts that the generator introduces: the traffic lights are not reconstructed, thus being predicted as anomalies.
     }
     \vspace{-20pt}
    \label{fig:generative}
 \end{figure}

{
% On the other hand, generating images with high fidelity is particularly complex in semantic segmentation due to their various content. Thus, generative approaches tend to create artifacts not only if a pixel of an unknown class is present, but also for pixels of known ones (see Fig.~\ref{fig:generative}).
On the other hand, generating images with high fidelity is particularly difficult in semantic segmentation due to their complex content. Thus, generative approaches tend to create artifacts not only when synthesizing pixels of unknown classes, but also pixels of known ones (see Fig.~\ref{fig:generative}).
}

{
% In this paper, we argue that it is fundamental to address the anomaly segmentation problem directly at the level of class scores: if we would have a model knowledge for each known class, we could determine if a pixel seen at test time should be considered or not an anomaly by measuring its distance with respect to such representations. We concretely pursue this intuition  by means of class-specific prototypes,  considering a pixel as anomalous whenever the highest matching score for the set of prototypes of the known classes is low.
In this paper, we argue that it is fundamental to address the anomaly segmentation problem directly at the level of class scores. Our intution is that if a model learns general but discriminative representation of each class, %can correctly represent %, by having a model knowledge for each known class, 
it can detect anomalies as pixels that are not compatible with any of the class representations. %class score of each test pixel by only measuring its similarity with respect to such representations and thus, using the score, deciding if it is anomalous or not. 
We concretely pursue this idea by means of class-specific prototypes, considering a pixel as anomalous whenever the highest matching score for the set of prototypes of the known classes is below a certain threshold.} We train the prototypes using a cosine classifier that bounds the class specific scores while ensuring that the prototypes embody the %the average 
average pixel features of the corresponding known class. It is important to emphasize that, to avoid the normalization problems of softmax-based approaches, we estimate the anomalies \textit{directly} from the compatibility between a feature vector and the prototypes. %, % of the prototypes \massi{preserving the original compatibility between a pixel and a known class}. %transforming them into probabilities.
%\dario{però poi trasportiamo tutto in range 0-1 e nel paragrafo 3.3 diciamo che usiamo probabilità} We call our approach PAnS, Prototypical Anomaly Segmentation.
We test our model PAnS on the popular StreetHazards benchmark \cite{hendrycks2019benchmark}, showing that it largely surpasses the state of the art. %performs better than expensive generative approaches while largely surpasses the state of the art. %An ablation study completes our experiments.

\myparagraph{Contributions.} To summarize, our contributions are:
\begin{itemize}
    \item We propose a novel perspective for the anomaly segmentation problem, revisiting the importance of class-specific scores rather than probabilities in the process of recognizing anomalous pixels.
      \item We present our approach, \textbf{P}rototypical \textbf{An}omaly \textbf{S}egmentation (PAnS), that computes such scores as the compatibility between a feature vector and class-specific prototypes, learning the latter as weights of a cosine classifier. %We name our approach . %new algorithm named PAnS, Prototypical Anomaly Segmentation, based on prototype learning.
   %We revisit the importance of class-specific scores rather than probabilities in the process of recognizing anomalous pixels.  learning class specific prototypes as weights of a cosine classifier. %through the classifier's weights. 
    \item Experiments on the widely adopted %public database for anomaly segmentation, \ie 
    StreetHazards showing that our approach surpasses the previous state of the art by a margin.
\end{itemize} 
%To address this problem, multiple works
% anomaly, previous works, difetti, noialtri, contributions.

% However, since the amount of semantic labels  to The amount of information sight can provide us is undoubtedly rich. As humans, the ability to extract this information allows us to \eg understand the current environment and take decisions, even simple ones, accordingly. Let us suppose we are driving a car and we see the scene captured in Fig. \ref{fig:teaser}. Recognizing that a pedestrian is crossing the road forces us to react properly by, \eg braking and avoiding an accident. We can take this life-saving decision because our visual system allows us to precisely recognize what is present in our environment and where it is present. In the path toward autonomous agents navigating in the real-world is crucial to equip our machines (\eg autonomous cars) with visual models with the same capabilities of recognizing what is present in an image and where it is present. Under this perspective, we cannot simply extract coarse information, but we would like to equip our visual models with the ability to infer the semantic content of an image at its finest level. % rely on coarse-level annotations (e.g. object classification \cite{alex})

\section{Related works}
\vspace{-5pt}
\label{sec:related}
In this section we review the topics that constitute the building blocks of our work, \ie uncertainty estimation in semantic segmentation, out-of-distribution (OOD) detection, anomaly segmentation and prototype learning. 

\myparagraph{Semantic segmentation.} %\dario{Rephrase - this copy and paste from Fabio's paper}
Modern semantic segmentation architectures \cite{long2015fully, chen2018encoder, zhao2017pyramid, lin2017refinenet, zhang2018exfuse} %have consistently pushed forward progress in semantic segmentation. 
%The state-of-the-art approaches 
are fully-convolutional encoder-decoder networks \cite{long2015fully, badrinarayanan2017segnet} that differ on the strategy used  to integrate contextual information in the pixel-level features. We can categorize these works as belonging mainly to two different approaches: 
%There are two main approaches: 
pyramid-based approaches \cite{zhao2017pyramid,lin2017refinenet,chen2017rethinking, chen2017deeplab, zhang2018exfuse, chen2018encoder}, that integrate modules exploiting information at different scales, and attention-based approaches \cite{yu2020context, yuan2019object, yu2018learning, yuan2018ocnet, fu2019dual, zhang2018context} that aggregates the long-range spacial dependencies using attention modules at different levels.
A drawback shared by all these architectures is that they require huge amount of training data, which is often time consuming and extremely expensive to collect. Moreover, they only consider an offline setting, \ie once the model has been trained, it is not possible to integrate additional knowledge. Although recent works have tried to move forward and deal with the addition of novel classes \cite{michieli2019incremental,cermelli2020modeling}, none of these approaches deal with anomaly detection.

%\massi{Questa parte \'e brutta. Dobbiamo stressare le differenze con gli altri, anche con ODIN (vedi R3)}
\myparagraph{Out of distribution detection} is a topic that has aroused growing interest in the machine learning community in recent years \cite{hendrycks2016baseline,liang2018odin,hsu2020generalized}.

\cite{hendrycks2016baseline} has established the standard baseline for out-of-distribution (OOD) detection where a threshold applied on the maximum softmax probability (MSP) is used to recognize whether a sample belongs to the training distribution (in-distribution) or not (out-of-distribution). % \new{by comparing the MSP with a certain threshold}. 
Beside its simplicity and effectiveness, the largest softmax probability is sub-optimal to detect anomalies %as a high confidence status of the model mainly 
for two reasons. 
% Firstly, it produces high confidence values even when the prediction is wrong, making it hard to distinguish it from the correct ones \cite{corbiere2019addressing}. 
%  Moreover, the high probability produced by the model reflects a very high confident status. Consequently, in the case of misclassification errors, the model assigns them low probabilities, which may be confused with the ones it should be assigned to OOD samples, as also for them the confidence of the model should be very low.
First, the model may produce high probability values even when the predictions are incorrect \cite{corbiere2019addressing}. Second %In this case, even if the model is failing, it is considered highly confident \cite{corbiere2019addressing}. Moreover, 
in case of correct predictions with %the model may also assign them 
low probabilities, %. Pixels with low probabilities may be 
the model may misinterpret them as OOD samples. %, as the model's confidence should be very low for them as well.}

% come l'ha scritto ECCV20
% measured the agreement between the classifier and a modified nearest-neighbor classifier on the test examples as a confidence score.
Other approaches have attempted to mitigate these issues. \cite{jiang2018trust} preprocesses the training set in order to find class-specific set of images %\massi{NC 
%\new{\dario{non so scriverla meglio. I set sono per ogni classe un cluster con solo i punti più simili, densi, toglie quindi gli outlier} high-density-sets for each class. For each class, these sets 
that contain only the training samples predicted with high similarity, %belonging to the given class class with high density, 
filtering out the fraction of sparse samples that may be outliers.  Given a test sample, the model detects anomalies %\massi{NC da ridurre
% as similar to a modified nearest-neighbor classifier, they define the score as the ratio of two values: at the numerator they compute the distance between the test sample and the nearest class different from the predicted class, while at the denominator they compute the distance of the sample from the predicted class.
through a modified nearest-neighbor classifier, computing the prediction as a ratio of distances between class sets. In \cite{corbiere2019addressing}, an additional neural network is trained on in-distribution data to produce high confidence values when the prediction of the model is correct. This additional architecture helps to model the confidence on the network predictions. % if the model prediction is trustworthy or not. 

Previous approaches \cite{gal2016dropout, kendall2017uncertainties} have also used Monte Carlo Dropout (MC-Dropout) for Bayesian approximation. With multiple forward pass, they compute the variance and the entropy as a measure of uncertainty.

ODIN \cite{liang2018enhancing} improves MSP by introducing a temperature scaling factor in the softmax operation and a small perturbation on the features before the classification step. The temperature value and the perturbation are computed on a OOD validation set. It is worth noticing that despite the apparent similarity, our work is conceptually different from \cite{liang2018enhancing}. First, \cite{liang2018enhancing} relies on a softmax function with temperature, while we directly compute the class scores as compatibility with the prototypes. Second, while \cite{liang2018enhancing} requires an OOD set, we train our model on in-distribution data only. % to  the scores before divide the scores provided by the network before converting them into probability by means of the softmax function. Using the cosine classifier, instead, the classifier weights and extracted features are divided by the norm of the respective vector \textit{before} calculating the dot product that will produce the scores.}

\myparagraph{Anomaly segmentation.} %Starting from OOD detection works, man \cite{lee2018simple, liang2017enhancing, lee2017training, hendrycks2018deep, devries2018learning, hsu2020generalized} have improved the ability of a model to recognize OOD examples from different points of view. However, all of these approaches focus on determining whether an entire image is an OOD example or not. In this paper, instead, we focus on the more difficult and realistic task where the model is asked to recognize each pixel of an image as in- or out-distribution, with the goal of segmenting anomalous regions from an image. This scenario is known as \textit{anomaly segmentation} (AS) \cite{hendrycks2019benchmark, xia2020synthesize, baur2018deep}. 
Current mainstream approaches for AS exploit pixel-wise reconstruction loss with auto-encoders (AEs) \cite{baur2018deep, haselmann2018anomaly}. The main disadvantage of these approaches is that when the training scene is very complex, as often is the case in roads or streets scenes, AEs are not able to model correctly the in-distribution and therefore they cannot guarantee to generate plausible in-distribution images from out-of-distribution regions. As a matter of fact, the CAOS benchmark \cite{hendrycks2019benchmark} demonstrated that MSP \cite{hendrycks2016baseline} outperforms AEs and Bayesian network-based approaches. Recent works \cite{lis2019detecting} and \cite{xia2020synthesize} proposed to compare test images containing OOD pixels with their reconstructed versions. In both cases, {images are reconstructed from the predicted semantic maps by means of a generative approach, \ie pix2pixHD \cite{wang2018high} and SPADE \cite{park2019semantic} respectively}. %To find the discrepancies between the two set of images, i.e. 
To detect the OOD pixels, \cite{lis2019detecting} and \cite{xia2020synthesize} measure the discrepancy between the original and reconstructed images by means of %respectively used 
a discrepancy network and a comparison module respectively. A drawback of these works is that artifacts in the reconstructions process may be wrongly recognized as anomalies, as shown in Fig.~\ref{fig:generative}.

Differently from these works, we do not rely on expensive generative approaches and we directly produce predictions from the compatibility between features and class prototypes. Our approach achieves better results despite being more lightweight and simple than generative-based ones. %While these two seminal works focus much more on anomalies rather than errors, our work detours from this conceptual approach to the problem and takes a completely different path: we focus more on on avoiding mistakes by considering in-distribution pixels as anomalies, purposefully avoiding to use a test time generator but opting instead of a lightweight class-specific prototype approach.

%takes a completely opposite path, not requiring a test time generator and focusing much more on avoiding mistakes by considering in-distribution pixels as anomalies.

\myparagraph{Prototype Learning} %is a classical method in representation learning literature. While multiple works focused on prototype learning on hand-crafted features (see \cite{liu2001evaluation for a comprehensive survey)}
\cite{yang2018robust, guerriero2018deepncm, mettes2019hyperspherical}, as opposed to softmax-based CNN, learn a metric space in which labeling is achieved by measuring the distance between the test image and prototypes of of class. 
Recently, prototype learning has been exploited by few-shot \cite{gidaris2018dynamic, qi2018low, snell2017prototypical, yang2018robust} and zero-shot \cite{xu2020attribute, liu2020convolutional} learning methods.
In this work, we take inspiration from \cite{gidaris2018dynamic, qi2018low}, using a cosine-similarity based classification layer that forces the classification weights to represent the class prototypes.

\begin{figure*}[ht]
    \centering
    \includegraphics[width=0.95\linewidth]{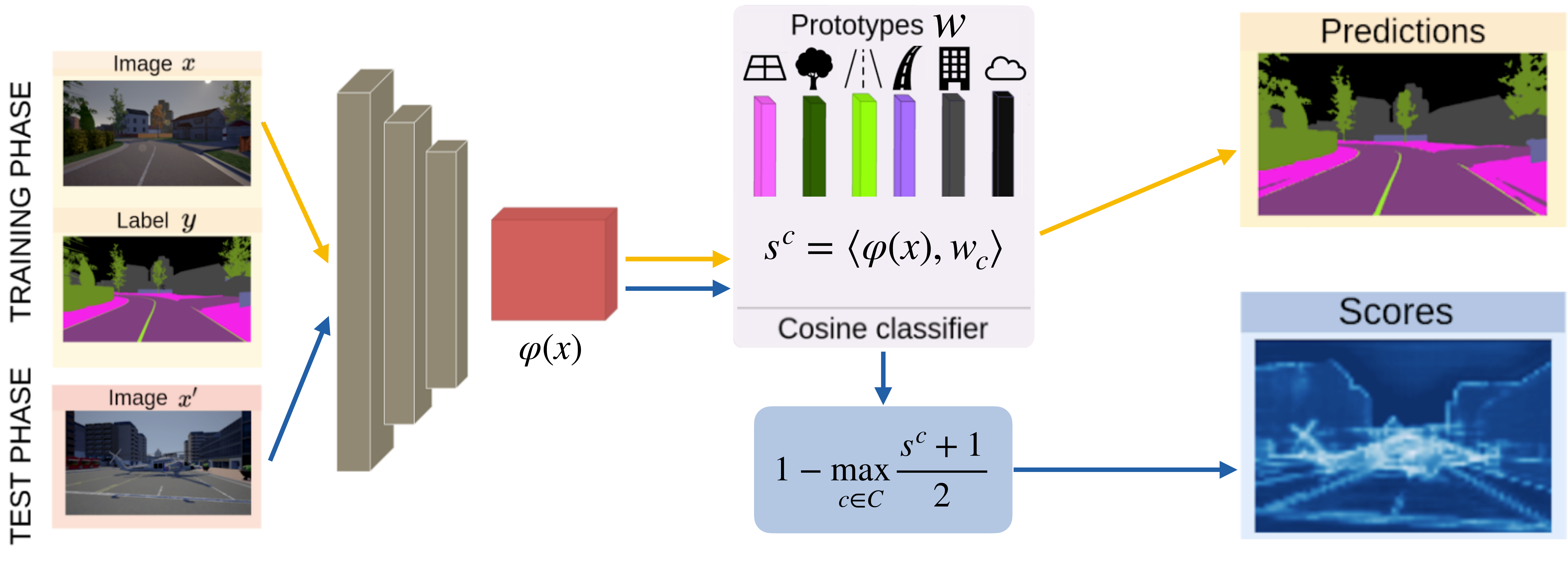}
    \vspace{-10pt}
     \caption{Overview of \ours. During training, we learn class-specific prototypes using a cosine classifier that computes the cosine similarity between the features $\varphi(x)$ and the classifier weights $w$. %The prototypes enable to compute a class-specific scores representing the confidence on each class, providing a reliable metric to derive the anomaly score.
     When a test image arrives, we compute the network prediction on the known classes and we compute the anomaly score by means of the cosine classifier scores.
     }
     \label{fig:method}
     \vspace{-15pt}
\end{figure*}

\vspace{-5pt}
\section{Anomaly Segmentation with Prototypes}
\vspace{-5pt}
\label{sec:method}
In this section we first formalize the anomaly segmentation problem (Section ~\ref{sec:problem}), highlighting the limitations of softmax-based approaches. We then describe how we overcome these limitations with our model, \expandednick~(\ours), in Section ~\ref{sec:ours}. %and how we overcome them with our method. and how will overcome them with our Prototythen we will focus on the limits that the previous approaches have, in order to highlight how simple while effective our solution solves the issues that no one had realized to be fundamental in anomaly segmentation.
An illustration of \ours\ is provided in Fig.~\ref{fig:method}.

%\subsection{Preliminaries}
\subsection{Problem formulation}
\label{sec:problem}
% Problem formulation
The goal of anomaly segmentation is to recognize which pixels in the image belong to anomalous objects, unseen during training. Let us denote as $\mathcal{X}\in\real^{|\mathcal{I}|}$ the image space, where $\mathcal{I}$ is the set of pixels. % with $w$ and $h$ the width and height of the image respectively. For the sake of clarity, we assume the images are composed by a set of pixel $\mathcal{I}$, with cardinality $|\set I| = w \times h$. 
During training, we are given a dataset $\mathcal{T}=\{(x_k,y_k)\}_{i=k}^{N}$ where $x\in\mathcal{X}$ is an image and $y\in\mathcal{Y}$ is its corresponding ground-truth mask. As in standard segmentation, $\mathcal{Y}$ contains pixel-level annotations for a set of semantic classes $\mathcal{C}$, \ie $\mathcal{Y}\in\mathcal{C}^{|\set I|}$. Given $\mathcal{T}$, we want to learn a function $f$ mapping an image to its corresponding anomaly score at pixel level, \ie  $f:\mathcal{X}\rightarrow \real^{|\set I|}$. %\fabio{Non vogliamo anche imparare le classi?} 
Without loss of generality, we consider $f$ built on three components. The first is a feature extractor $\varphi:\mathcal{X}\rightarrow\mathcal{Z}$ mapping images into a feature space $\mathcal{Z}\subset \real^{|\set I|\times d}$, with $d$ being the feature dimensions. The second is a scoring function $\rho:\mathcal{Z}\rightarrow\real^{|\set I|\times|\mathcal{C}|}$ mapping the features in $\mathcal{Z}$ to pixel-level class scores. The third is an anomaly score function $\sigma:\real^{|\set I|\times|\mathcal{C}|}\rightarrow \real^{|\set I|}$, mapping the class scores to the final anomaly ones. %prediction. %pixel into its corresponding in the  and $y$  define a belong to 
%Let us formalize the AS problem. Suppose we have an initial training set 
% $\mathcal{T}_{in}={(x_i,y_{i,c})}_{i=1^{N_0}}$, with $x_i$ being an image in the image space $\mathcal{X}$, $y_i$ being the groundtruth of each pixel in the image-label space $\mathcal{Y}_{in,C}$ in which $\mathcal{C}$ is the label set containing all the class categories found in $\mathcal{X}$. Then, suppose there exist another set $\mathcal{T}_{out}$ containing only unseen categories, i.e., $\mathcal{Y}_{in} \bigcap \mathcal{Y}_{out}=\emptyset\; \forall c \in [0,C-1]$, which we refer to as \textit{anomalies} in this work. As the model will be evaluated on a test set $\mathcal{T}_{test}$ = $\mathcal{Y}_{in} \bigcup \mathcal{Y}_{out}$. we aim at learning a function  $f:X\rightarrow \mathcal{Y}_{in,C} \bigcup \textit{a}$ able to map an image $x$ into either the set of learned semantic classes $\mathcal{Y}_{in,C}$ or the anomaly class $\textit{a}$. 
% \dario{%da riscriverre con altre parole, presa da owr
% Without loss of generality, we consider $f$ being built on three components: a feature extractor $\omega:\mathcal{X}\rightarrow\mathcal{Z}$ mapping images into a feature space $\mathcal{Z}$; a scoring function $\rho:\mathcal{Z}\rightarrow\Re^{|\mathcal{K}^T|}$ mapping features in $\mathcal{Z}$ to known class scores; and $\sigma:\Re^{|\mathcal{K}^T|}\rightarrow \mathcal{K}_t \cup \text{a}$, mapping the class scores to the final prediction.}
%The core of the anomaly 
A core component of every anomaly segmentation algorithm is $\sigma$, that produces the final anomaly scores. In the following we discuss how previous approaches instantiated the $\sigma$ function.

\myparagraph{Maximum Softmax Probability (MSP).}
One of the most popular and effective approaches for anomaly segmentation is Maximum Softmax Probability (MSP) \cite{hendrycks2016baseline}. The intuition behind MSP is that the anomaly score of a pixel should depend on the highest probability assigned to any of the known classes. Given an image $x$ and its pixel-level class scores $s=\rho(\varphi(x))$, MSP defines the anomaly score for pixel $i$, \ie $\sigma_i(s)$, as:
\begin{equation}
    \label{sec:msp}
    \sigma_i(s) = 1 - \max_{c\in \mathcal{C}} \frac{e^{s^c_i}}{\sum_{k\in\mathcal{C}} e^{s^k_i}}
\end{equation}
where $s^k_i=\rho_k^i(z)$ is the score for class $k$ in pixel $i$. % belonging to class class feature vector of $z$ at pixel $i$ and $\pho_c$
Note that anomaly scores are defined as the inverse of the maximum probability assigned to any known class in $\mathcal{C}$, with the probabilities computed through the softmax function.  
% The softmax function %, also known as normalized exponential function, 
% is the most commonly used function to convert the logits produced by a model into class probabilities. %For each score, it takes its exponents value and divide it by the sum of the exponents of all the scores so that the output vector adds up to 1, becoming a probability score. It can be summarized in the following formula:
% It is defined as:
% %%%% da togliere 
% \begin{center}
% $\sigma = \frac{e^{z_i}}{\sum_{j=1}^{k}e^{z_j}}$
% \end{center}
Despite its effectiveness, we argue that using softmax probabilities is not the best choice to estimate the anomaly scores. %Beside its wide usages in several different tasks, we argue that it is not the best 
%choice for anomaly segmentation. 
% dependent on the metric used. Vice versa, a pixel is considered as an in-distribution example when the confidence of the model is high enough to overcome the chosen threshold. 
% When using the softmax function, the confidence of the model prediction on each pixel gets smoothed, sometimes flattened, leading to very confident pixels being considered as uncertain after the softmax normalization.
% Using the softmax function, the confidence of the model prediction on each pixel gets smoothed, sometimes flattened, leading confident pixels with high predicted scores to be considered as uncertain.
In fact, the softmax function may smooth the confidence of the model prediction on each pixel, %(\eg when two classes are predicted with high scores), 
leading to consider uncertain (and thus anomalous) pixels even with high predicted initial scores. % scores to be uncertain.
% In this scenario, smoothing the probabilities is dangerous, since a pixel is considered  an \textit{anomaly} when the confidence of the model on that pixel is below a certain threshold. 
%This is problematic, since pixels with low confidence may be wrongly predicted as anomalous. %, using the softmax probabilities to detect anomalies is sub-optimal in this context since a pixel is considered anomalous when the confidence of the model in that pixel falls below a certain threshold.

As a toy example, suppose we have two different classes and, given a pixel, the model produces a very high score for the first class and a very low one for the other one. %scores for the second and the third one. 
% In this case, we will have a low-entropy in the probabilities after applying the softmax function, with a very high probability for the class with the highest score, and low probability for the other. However, if the two classes have both very high but close scores, both of them will receive a similar probability value after the softmax normalization, with a high entropy on the prediction for the corresponding pixel.
In this case, the probabilities would have a low entropy after applying the softmax function, with a high probability for the class with the highest score and a low probability for the other. The model will correctly consider this pixel as not anomalous. However, if both classes have high but near scores, they will get close probability values after softmax normalization. This high entropy on the pixel prediction indicates that the model is uncertain of the semantic of the pixel, but the high initial scores may hint that the model \textit{is not uncertain} that the pixel belongs to a known class. % prediction for the pixel, thus the cha.

Note that we will get this high entropy class scores regardless of the initial magnitude of the scores. Obviously, since the magnitude of the logits before the softmax are unbounded, it is not trivial to assess whether the model is uncertain on the semantic content due to the pixels featues being out of distribution or not. %considers pixels to be  of each score before normalization. 
In the following, we will show that keeping the class scores independent and constraining them to be bounded in a known range (\ie $[-1,1]$) through prototype matching is useful to assess the actual confidence of a model in its predictions, improving the recognition of anomalous pixels. %predictio As we  for two of the classes, both of h. In fact, it will produce two similar probability values for both the first and the second scores which will only be slightly greater than the probability of the third one, thus reducing the significant gap that there was between the scores instead. This is not a desirable situation because the low probability that could be assigned to high scores could be below the metric-dependent threshold even if the network is very confident on those pixels. This is particularly significant for the pixels on the boundaries, since the network could be found to be very confident on both the classes divided by the boundary pixels.

\subsection{\expandednick}
\label{sec:ours}
% In the previous section, we have analyzed how %the problems of previous approaches, we argue that they fail in evaluating the confidence of the model on the anomalies. In fact, 
% MSP fails in evaluating the likeliness of an object using the probability after the softmax normalization, which discards information about the model confidence. %On the other hand, the generative approaches, comparing the original image and its reconstruction are prone to introduce generation errors which lead to detect non-anomaly pixels as an anomalous one. 
% For this reason, we take a different approach and we argue that it is important to consider the confidence on each class independently. %they both fail in effectively evaluating the confidence of the model  they are not considering the similarity of a pixel with class
% %What we really aim to find anomalies is to find pixels which are far from the model knowledge. 

% In the previous section, we have analyzed how MSP fails in evaluating the likeliness of an object using the probability after the softmax normalization, which discards information about the model confidence. For this reason, we take a different approach and we argue that it is important to consider the confidence on each class independently.
In the previous section, we discussed how MSP may fail in recognizing anomalous pixels  %the  of an object using probability after 
due to the softmax normalization, which discards information about the confidence of the model. As a consequence, we take a different approach, arguing that it is critical to consider the confidence of each class separately.
Ideally, we want to obtain confidence values which: (i) are independent for each class, (ii) do not require extra-computation, and (iii) are bounded in a certain range, such that it is possible to define a threshold on their scores to detect anomalies.
% \dario{opposto a quello che diciamo nell'intro i.e. non emettiamo una probabilità}.

{
% In this work, we argue that an effective solution to accomplish this is to represent each class using a prototype. In fact, having a prototype for each class allows to produce a confidence score independently for each class, computing how close are the features of any pixel to it. 
% Different ways to define class prototypes can be found in the literature \cite{guerriero2018deepncm, gidaris2018dynamic, qi2018low, yang2018robust, snell2017prototypical}.
% The easiest solution would be computing the class prototypes by means of averaging all the visual features for a given class and employing the Euclidean distance between the features and the prototypes to compute the confidence score. However, this solution does not satisfy our desires since it is computationally expensive to compute the average of the features; also, the confidence scores would not be bounded in a pre-defined range. 

To accomplish this, we propose to represent each class using a prototype. Each class prototype may be considered as a reference feature vector for a certain class.  %scores represent the compatibility between a feature vector and the class prototype. %In fact, having a prototype for each class allows to
We can then compute class-independent confidence scores by computing the similarity between the features of any pixel and the prototype itself. 
%In literature can be found multiple ways to define the class prototypes \cite{guerriero2018deepncm, gidaris2018dynamic, qi2018low, yang2018robust, snell2017prototypical}. 
%However, some solutions do not satisfy our goals, being either computationally expensive or producing an unbounded similarity. As an example, DeepNCM \cite{guerriero2018deepncm} computes the class prototypes by means of averaging all the visual features for a given class, which may be expensive, and computes the similarity between the features and the prototypes making use of the Euclidean distance, which is unbounded.
% Among the different ways to define class prototypes that can be found in the literature \cite{guerriero2018deepncm, gidaris2018dynamic, qi2018low, yang2018robust, snell2017prototypical}, 
We take inspiration from few-shot classification learning works \cite{qi2018low, gidaris2018dynamic}, and we use a simple yet effective cosine classifier, which implicitly encodes the class-prototypes by means of its classification weights.
}
%we argue that an effective way to extract the prototypes is using a cosine similarity-based classifier. 
%To do so, we argue that it is important to extract a class-prototype from the training data, such that, when testing data arrive, we are able to evaluate how similar are the pixel features with the class prototype. In this way, we are able to effectively recognize if a pixel belong to a certain class, or if it is dissimilar to all the known classes, \ie it is an anomaly.
%\fabio{Dire perché non si può usare NCM, ovvero perché abbiamo bisogno di score normalizzati per calcolare l'anomaly score.}
%Having analyzed the problems that the use of the softmax function brings, our work is based on the intuition that it is not the probability value produced by the model that indicates its confidence on each pixel, but the actual scores on which this probability is computed. In fact, although the softmax function flattens the values of the scores in a range of probabilities, in origin these are much more informative. The reason behind this behavior is intrinsic to the structure of the softmax function itself, i.e. to provide the probability value of a score it must take into account all the other scores, flattening the entire probabilities output vector. It would be much more desirable, instead, to keep the disparities between values as much as possible, so that when the predicted value is low it could be due to an anomaly detection and not to a prediction artifact. Fig. \ref{fig:qualitative} visually supports our claims.

\myparagraph{Cosine Classifier}
To effectively extract class-prototypes from the network, we use a cosine classifier, \ie a classifier which uses cosine similarity between the input features and the class weights as class scores. %-based classifier, in short cosine classifier. 
While this classifier has been used for image classification \cite{gidaris2018dynamic, luo2018cosine, NIPS2016_90e13578, Hou_2019_CVPR} to effectively learn class-prototypes, we are the first to use this classifier with the aim of recognizing anomalies in semantic segmentation.
%Since semantic segmentation is a pixel-wise classification problem, 
We %adapt the cosine classifier from classification to segmentation by 
replace the standard convolutional classifier, with a cosine similarity-based one. %fully-convolutional classifier in place of the standard convolutional one. To do so, we need to normalize the classificator's weights $w$ and the features at each pixel, before computing the per pixel class logits. %the convolutional operation.
In particular, the classification scores for a class $c$, given an image $x$ and a pixel $i$, are:
\begin{equation} \label{eq:cosine_class}
    s^c_i = \rho_c^i(\varphi(x)) =   \langle \varphi_i(x), w_c\rangle = \frac{\varphi_i(x)^\intercal w_c}{||\varphi_i(x)||\;||w_c||},
 \end{equation}
where $\varphi_i(x)$ is the output of the feature extractor $\varphi$ at pixel $i$ of the image $x$, and $w_c \in \real^d$ is the prototype of class $c$.
We note that the scores $s$ are in the range $[-1, 1]$ due to the normalization effect on the denominator.

To learn the prototypes, % the cosine classifier is similar to training a linear classifier. In fact, 
we apply the standard cross-entropy loss on the softmaxed probabilities computed on the scores $s^c$:
\begin{equation}
    \ell_{CE}(x, y) = - \frac{1}{|\set I|} \sum_{i \in \set I} \log \frac{e^{\tau s^{y_i}_i}}{\sum_{c \in \mathcal{C}} e^{\tau s^c_i}},
\end{equation}
where $\tau$ is a scalar value that scales the classification scores in the range $[-\tau, \tau]$ and $y_i$ is the ground-truth at pixel $i$. 
Intuitively, minimizing the $\ell_{CE}$ loss forces the prototype weights to have low cosine distance with features of their respective class, representing them on average. 
%Furthermore, since the feature vectors belonging to a class must are pushed to be close to the prototype of that class, the features extractor $\varphi$ is forced to generate $L_2$ normalized features vectors with very low intra-class variance. This further improves the networks in being more confident when it encounters pixels of known classes and much less confident with anomaly pixels.

\myparagraph{Computing the Anomaly Scores}
%After defining the prot, we now introduce how to effectively 
In this section, we describe how we use the prototypes to detect the anomalies in the input image. To overcome the limitations of softmax function, we argue that it is important to avoid the use of normalized probabilities, but rather to use directly the classification scores.
Since the cosine classifier outputs the similarity of each weight with the visual features extracted from the network, it enables the use of the scores $s$ as a confidence measure on the presence of a class.
Moreover, we can exploit the fact that the scores of the classifier are bounded in the range $[-1, 1]$ and define the binary probability $\bar{\sigma}$ of a class $c$ to appear in a pixel $i$ of an image $x$ as:
\begin{equation}
    \bar{s}^c_{i} = \frac{s^c_i + 1}{2}
\end{equation}
and the anomaly score $\sigma$ using the maximal binary probability, which we define as:
\begin{equation}
    \sigma_{i}(s_{i}) = 1 - \max_{c \in \set C} \bar{s}^{c}_i.
\end{equation}
Intuitively, $\sigma_i$ will produce scores close to $1$ when the visual features are far from all the class-prototypes, while
close to $0$ if at least the prototype of one class is close to them.

With this strategy, we expect that our method is able to effectively represent the known classes, having an high confidence on pixels belonging to them, while we expect no class prototype to be close to the extracted features for pixels of anomalous objects. %eon pixels belonging to an anomaly object we expect that no class prototype is close to the extracted features, obtaining an high anomaly score. 
Moreover, we avoid the aforementioned issues of the softmax function largely boosting the results as we will demonstrate in the experimental section.

\section{Experiments}
\vspace{-5pt}
\label{sec:experiments}

\begin{figure}[t]
    \vspace{5pt}
    \centering
    \includegraphics[width=0.9\linewidth]{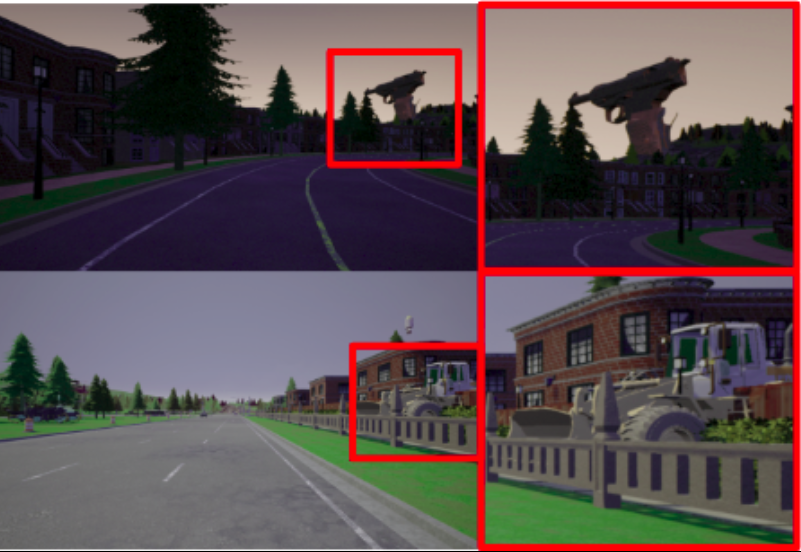}
     \vspace{-6pt}\caption{Examples taken from StreetHazards dataset \cite{hendrycks2019benchmark}.
     \vspace{-14pt}
     }
    \label{fig:dataset}
 \end{figure}

\myparagraph{Dataset and baselines}
We conduct our experiments on the popular StreetHazards \cite{hendrycks2019benchmark} database, which has been proposed within CAOS benchmark \cite{hendrycks2019benchmark} as a synthetic dataset for anomaly segmentation. It contains 5125 training images with paired semantic labels, 1031 validation images without anomalies and 1500 test images with anomalies. As the authors used Unreal Engine along with CARLA simulator \cite{dosovitskiy1711carla} to obtain the synthetic images, they selected different towns for the three splits. Test images contain anomalies randomly selected from a set of 250 objects. These objects are placed in the test images trying to reproduce plausible road scenarios. Figure \ref{fig:dataset} shows examples of images taken from the dataset.

On this benchmark we compare our method with state-of-the-art anomaly segmentation approaches, namely MSP \cite{hendrycks2016baseline}, MSP + CRF \cite{hendrycks2019benchmark}, an auto-encoder (AE) based approach \cite{baur2018deep}, Dropout \cite{gal2016dropout}, and the generative approach SynthCP \cite{xia2020synthesize}. 

% In \textbf{MSP} \cite{hendrycks2016baseline}, after computing the scores for each pixel, the model produces the probability output vector by computing the softmax function over the scores. Computing the maximum for each pixel, both known and anomaly ones, it then obtain the maximum probability that each pixel belongs to one of the known classes.

\myparagraph{Metrics.} Following \cite{hendrycks2019benchmark, xia2020synthesize, lis2019detecting} we used as anomaly segmentation metrics AUPR, AUROC and FPR95 as they are widely used in out-of-distribution detection \cite{hsu2020generalized, liang2017enhancing}. AUPR measures the area under the Precision-Recall curve, AUROC measures the area under the TRP and FPR and FPR95 measures the FPR at 95\% of recall. In each of these metrics, anomaly pixels have been considered as positives and all the others as negative.

\vspace{-5pt}
\subsection{Implementation details}
\vspace{-5pt}
%\dario{SynthCP dicono nel paper che hanno usato resnet101 e nel codice 50 come noi}. 
Following \cite{xia2020synthesize}, we used a ResNet-50 architecture \cite{he2016deep} as a backbone and PSPNet \cite{zhao2017pyramid} as the head module. We trained our segmentation module for 40 epochs, with batch size equal to 2 and a learning rate equal to 0.007. We used as learning rate decay policy the polynomial schedule with power equal to 0.9 and a weight decay equal to 0.0001. We also used InPlace-ABN \cite{rotabulo2017place} which allows to save up to 50\% of GPUs memory. Same as \cite{xia2020synthesize}, we used multiple scale evaluation at test time and at training time we performed random scale, random crop and random horizontal flip augmentation.

%\dario{We report in Fig. \ref{fig:generative} the results of the synthetic images.}

\begin{table}[t]
\centering
\resizebox{\linewidth}{!}{
\begin{tabular}{l|ccc}
Method & \multicolumn{1}{c}{AUPR $\uparrow$} & \multicolumn{1}{c}{AUROC $\uparrow$} & \multicolumn{1}{c}{FPR95 $\downarrow$} \\ \hline
AE \cite{baur2018deep} & 2.2 & 66.1 & 91.7 \\
Dropout \cite{gal2016dropout} & 7.5 & 69.9 & 79.4 \\
MSP \cite{hendrycks2016baseline} & 6.6 & 87.7 & 33.7 \\
MSP + CRF \cite{hendrycks2019benchmark} & 6.5 & 88.1 & 29.9 \\
SynthCP \cite{xia2020synthesize} & \textbf{9.3} & 88.5 & 28.4 \\ \hline
\textbf{\ours} & 8.8 & \textbf{91.1} & \textbf{23.2}
\end{tabular}}
\vspace{-8pt}
\caption{Results on StreetHazards dataset \cite{hendrycks2019benchmark} according to AUPR, AUROC and FPR95 metrics. \cite{hendrycks2019benchmark}.
\vspace{-10pt}
}
\label{tab:streethazards}
\end{table}

\begin{figure}[t]
    \centering
    \includegraphics[width=\linewidth]{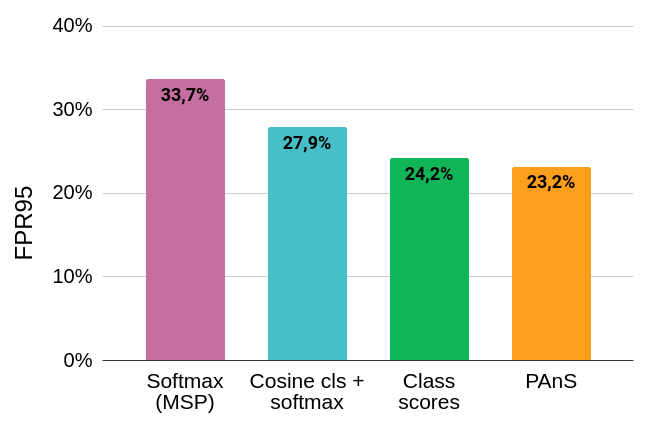}
    \vspace{-30pt}
     \caption{Difference between the direct usage of scores (both the scores produced by a standard classifier and a cosine-based one) on StreetHazard dataset \cite{hendrycks2019benchmark}.}
     \vspace{-15pt}
    \label{fig:ablation}
 \end{figure}

\begin{figure*}[t]
    \centering
    \includegraphics[width=0.94\linewidth]{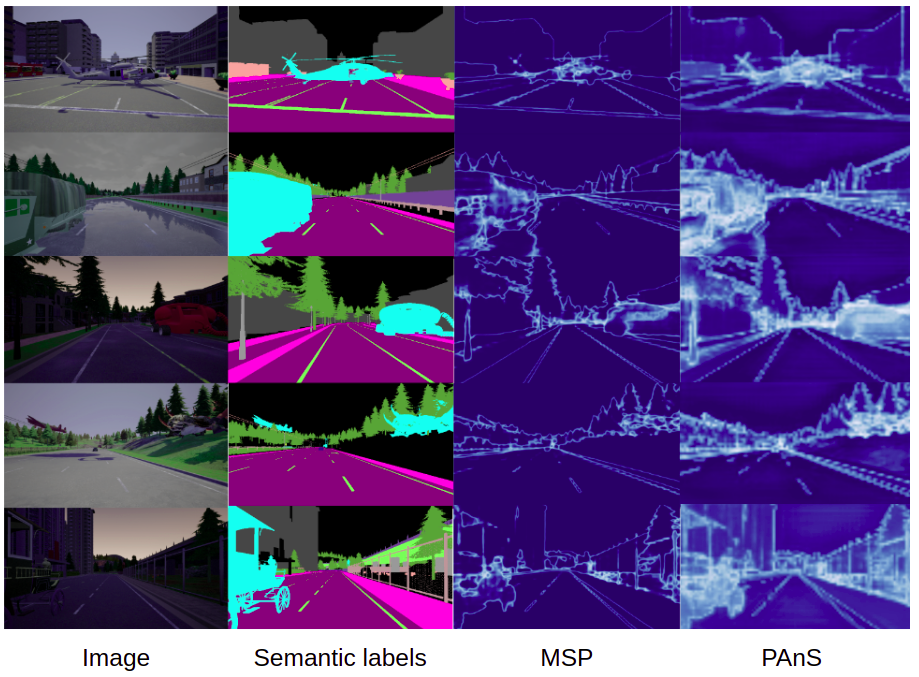}
    \vspace{-5pt}
     \caption{\textbf{Qualitative comparison} between the use of probabilities (MSP) and our direct scores (PAnS) for segmenting anomalies on StreetHazards \cite{hendrycks2019benchmark}. White indicates an high score for the anomaly, while the blue indicates a low score. In the semantic labels the anomaly are represented in cyan.}
    \vspace{-5pt}
    \label{fig:qualitative}
 \end{figure*}
 
\subsection{Comparison with the state of the art}
\vspace{-5pt}
The results of our comparison with the state of the art are reported in Table \ref{tab:streethazards}. As the table shows, in this scenario our method achieves the best performances by a margin, under AUROC and FPR95 OOD metrics, being instead comparable under AUPR values. Among all, noteworthy is the result of 23.2\% achieved under the FPR95 metric which shows that our method is much less prone to confuse pixels of known classes as anomalies. This is due to the fact that our prototype-based classifier betters preserve the original scores for known classes which might be either smoothed by the softmax normalization (as in MSP) or overwritten by inaccurate generations (as in SynthCP). %pixel of known as anomalies than the other methods. 
Indeed, under the FPR95 metric, our approach improves the state of the art (SynthCP) by almost +6\%. The fact that our prototypes induce a classifier more robust against misclassification of known class pixels is reflected in the other metrics. % , having a prototype for each known class encoded in the classifier allows the model to be more confident about them, 
Our approach reaches 91.1\% of AUROC, improving over the best previous method, SynthCP, of 2.6\%. These results confirm how our approach achieves the best trade-off between recognizing anomalous pixels while preserving high confidence predictions for pixels of known classes.
On the other hand, SynthCP obtains a slightly better AUPR (+0.5 w.r.t. \ours). However, we highlight that \ours\ only requires a single forward pass on the network, without any generative step and without increasing the computation required by the model. 

Note also that generative models might be affected by the quality of the generated images. Indeed, using synthetic images often introduces artifacts, as mentioned in \cite{xia2020synthesize}. These artifacts might hamper the performance of generative anomaly segmentation models, since they might be wrongly segmented as anomalies (\ie see Fig.~\ref{fig:generative}).  %introduce in AS errors because they are considered anomalies even when in reality they are not   % and without any post-processing on the scores %proves that superiority of our approach over the softmax-based ones. 

% \begin{figure}[t]
%     \vspace{5pt}
%     \centering
%     \includegraphics[width=0.9\linewidth]{images/generative_models2.png}
%      \caption{Qualitative results of SPADE \cite{park2019semantic} reconstructions on StreetHazards dataset \cite{hendrycks2019benchmark}. The green bounding box identifies a good reconstruction since it has not been corrupted by the anomaly. The red bounding box, instead, identifies the artifacts that the generator introduces. Indeed, the traffic lights are not reconstructed, thus being predicted as anomalies.
%      \vspace{-5pt}
%      }
%     \label{fig:generative}
%  \end{figure}
 
% \begin{table}[t]
% \setlength{\tabcolsep}{18pt} % Default value: 6pt
% \resizebox{\linewidth}{!}{
% \begin{tabular}{l|c}
% Method & \multicolumn{1}{l}{FPR95 $\downarrow$} \\ \hline
% Softmax (MSP \cite{hendrycks2016baseline}) & 33.7 \\
% Cosine cls + softmax & 27.9 \\
% Class scores & 24.2 \\
% Cosine cls + scores & \textbf{23.2}
% \end{tabular}
% }
% \caption{Difference between the direct usage of scores (both the scores produced by a standard classifier and a cosine-based one) on StreetHazard dataset \cite{hendrycks2019benchmark}.
% %\vspace{-8pt}
% }
% \label{tab:ablation}
% \end{table}

\begin{table*}[t]
\centering
\renewcommand{\arraystretch}{1.5} % Default value: 1
\setlength{\tabcolsep}{4pt} % Default value: 6pt
\resizebox{\linewidth}{!}{
\begin{tabular}{l|ccccccccccc|c}
%dario: standard to be fixed
Classifier & bkg & building & fence & pole & street-line & road & sidewalk & veget. & car & wall & t.sign & mIoU \\ \hline

Standard & 84.5 & 70.9 & 30.1 & \textbf{23.6} & \textbf{26.7} & 92.1 & 57.4 & 75.1 & 53.3 & 42.9 & 28.9 & 53.2 \\
Cosine-based & \textbf{84.8} & \textbf{72.1} & \textbf{30.9} & 22.3 & \textbf{26.7} & \textbf{92.5} & \textbf{60.0 }& \textbf{75.3} & \textbf{55.2} & \textbf{45.7} & \textbf{30.3} & \textbf{54.2}
\end{tabular}
}
\vspace{-10pt}
\caption{Comparison on IoU using a standard or the cosine classifier.} \label{tab:miou}
\vspace{-10pt}
\end{table*}

\myparagraph{Qualitative results} To analyze the impact of our cosine classifier and scores, we report in Fig. \ref{fig:qualitative} some qualitative examples on anomaly scores produced by the softmax-based approach MSP \cite{hendrycks2016baseline} and our approach \ours\ for randomly selected samples of StreetHazards. %cosine scores (PAnS). 
In the figure, white regions indicate higher anomaly scores while blue regions lower ones. As the figure shows, our model is able to correctly assign low scores to the regions where the anomaly is present (\eg \textit{helicopter} on the top image, \ie \textit{carriage} in the bottom), while MSP does not, covering only small portions of the anomalies. At the same time, both MSP and our approach tend to assign high anomaly scores to boundaries between known classes (\eg \textit{street lines} and \textit{road} in the top, \textit{building} and \textit{sidewalk} in the bottom). We believe modeling this highly uncertain regions between two or more known classes, is an open problem for anomaly segmentation algorithms,  which would be important to address in future works. 

\myparagraph{Ablation study of anomaly scores.} {We report in Fig.~\ref{fig:ablation} an ablation study about the choice of the anomaly score function $\sigma$ on the StreetHazards dataset \cite{hendrycks2019benchmark}. To provide a comprehensive comparison, we considered four variants: the softmax predictions of a standard linear classifier (MSP \cite{hendrycks2016baseline}), the softmaxed predictions of a cosine classifier (\textit{Cosine cls + softmax}), using the unnormalized class scores of a linear classifier (\textit{Class scores}), and finally the unnormalized cosine scores of \ours. % with the exploitation of the direct scores produced by the network both for a standard and prototype-based classifier.
As the figure shows, computing the softmax probabilities using a cosine classifier improves the performances w.r.t. making use of a standard classifier, improving the FPR95 value reached by the latter by 5.8\% points.
However, we note that using directly the class scores of the network instead of the softmax-normalized probabilities is highly beneficial, improving the standard softmaxed version by 9.5\% and the cosine one by 4.7\%. % or not (9.5\%).
%Moreover, we note that the cosine classifier improves the performance of MSP by 9.5\%

Finally, we note using the unnormalized cosine scores (\ours) outperforms the use of standard class scores achieving the highest FRP95 value up to 23.2\%. We ascribe this improvement to %However, we recall that using the direct scores of a standard classification later does not suffice our goal, since the network output is 
the unbounded nature of scores of a standard classification layer, where defining a threshold values for detecting anomalies is more difficult.} % fifficult to define a threshold.} % difficult to define a value for the anomaly threshold.}

\myparagraph{Ablation study of classifiers.} While our model \ours\ shows promising results on anomaly segmentation, an open question is whether it still maintains the high discrimination capabilities of standard classification modules.
In Table \ref{tab:miou} we report the IoU achieved by both a standard and a cosine similarity-based classifier on each class of the StreetHazards dataset. Overall, the cosine-based classifier outperforms the standard classifier obtaining a 54.2\% mIoU, which is 1\% more than standard classifier (53.2\% mIoU).
Results show that the cosine similarity allows the model to reach higher performances on almost every class, especially on classes that are frequently considered hard, such as fence, traffic sign, and sidewalk. %Remarkably, the model achieves a high IoU on small classes, as traffic sign and sidewalk, reaching respectively 30.3\% and 60.0\% of IoU. 
Only in the \textit{pole} category the model's performance slightly deteriorates w.r.t. the standard classifier, 22.3\% against 23.6\%. The reason is that in StreetHazards the \textit{pole} class is small and rarely represented, making it difficult for the model to estimate a good prototype for it. % Overall, the cosine classifier surpasses the traditional one also on meanIoU, reaching up to 54.2\% points against 53.2\%.

\section{Conclusions}
\vspace{-5pt}
\label{sec:conclusions}
In this paper we addressed the problem of anomaly detection in semantic segmentation, an important but yet scarcely researched topic. Previous works either focused on how to model the probability of a pixel belonging to an unknown class, or on generative approaches for detecting anomalies through reconstruction errors. %in training images through artificial artifacts. 
Here, we instead argued that to detect if a pixel is anomalous or not, it is more important to have a model that measures the distance between visual features extracted from a pixel and a general representation of each class, % prototypical of each class and then measure how far each test pixel is from this representation, 
rather than using maximum softmax probabilities. We obtain such general representations by learning class-specific prototypes %Our algorithm learns prototypical representations for each class through the use of 
as weights of a cosine similarity-based classifier. Experiments on the widely used StreetHazards benchmark support our intuition, as we achieve the new state of the art with a significant margin over previous work in two out of three metrics.

Future work will further explore this research avenue, looking into distinguishing between anomalous pixels and pixels at the boundary between two known classes, that are challenging to predict due to their uncertain nature.

\myparagraph{Acknowledgments} \newline
This work has been partially funded by the ERC 853489 - DEXIM, by the ERC grant 637076 - RoboExNovo and by the DFG – EXC number 2064/1 – Project number 390727645. Computational resources were provided by HPC@POLITO (http://www.hpc.polito.it)

{\small
\bibliographystyle{ieee_fullname}
\bibliography{egbib}
}

\end{document}